\documentclass[11pt]{article}
\usepackage[most]{tcolorbox}

\newtcolorbox{takeaway}[2][]{
  colback=gray!3,
  colframe=gray!50,
  boxrule=0.8pt,
  arc=1pt,
  left=3pt,
  right=3pt,
  top=2pt,
  bottom=2pt,
  coltitle=black,             
  title=\textbf{#2},       
  fonttitle=\bfseries,     
  arc=2mm,
  boxrule=1pt,  
  left=4pt, right=4pt, top=4pt, bottom=4pt, %
  #1                       
}
\usepackage[final]{acl}

\usepackage{times}
\usepackage{latexsym}

\usepackage[T1]{fontenc}

\usepackage[utf8]{inputenc}

\usepackage{microtype}
\usepackage{xspace}
\usepackage{inconsolata}
\usepackage{enumitem}
\usepackage{amsmath,amsthm,amssymb}
\usepackage{algorithm}
\usepackage{algorithmic}
\usepackage{graphicx}
\usepackage{colortbl}
\usepackage{dsfont}
\usepackage{float} 
\usepackage{caption}
\usepackage{booktabs}
\usepackage{makecell}
\usepackage{diagbox}
\usepackage{fancybox}
\usepackage{bm}
\usepackage{color}

\definecolor{LightBlue}{rgb}{0.85,0.92,1.0}

\newcommand{\ours}{MGI\xspace}

\definecolor{LightGray}{RGB}{235,235,235}
\usepackage{graphicx}
\usepackage{multirow}
\title{Why Multimodal In-Context Learning Lags Behind? \\ Unveiling the Inner Mechanisms and Bottlenecks} 

\author{
 \textbf{Yu Wang},
 \textbf{Sharon Li}
\\
Department of Computer Sciences, University of Wisconsin-Madison,
\\
  \texttt{\{yuwang, sharonli\}}@cs.wisc.edu
}

\begin{document}
\maketitle

\begin{abstract}
In-context learning (ICL) enables models to adapt to new tasks via inference-time demonstrations. Despite its success in large language models, the extension of ICL to multimodal settings remains poorly understood in terms of its internal mechanisms and how it differs from text-only ICL.
In this work, we conduct a systematic analysis of ICL in multimodal large language models.
Using identical task formulations across modalities, we show that multimodal ICL performs comparably to text-only ICL in zero-shot settings but degrades significantly under few-shot demonstrations. To understand this gap, we decompose multimodal ICL into task mapping construction and task mapping transfer, and analyze how models establish cross-modal task mappings, and transfer them to query samples across layers.
Our analysis reveals that current models lack reasoning-level alignment between visual and textual representations, and fail to reliably transfer learned task mappings to queries. Guided by these findings, we further propose a simple inference-stage enhancement method that reinforces  task mapping transfer. Our results provide new insights into the mechanisms and limitations of multimodal ICL and suggest directions for more effective multimodal adaptation. Our code is available  
\href{https://github.com/deeplearning-wisc/Multimocal-ICL-Analysis-Framework-MGI}{here}.
\end{abstract}
\section{Introduction}
In-context learning (ICL) enables models to adapt to new tasks at inference time by using a small number of demonstrations, without updating parameters~\cite{zhang2023makes,min2022rethinkingroledemonstrationsmakes,zhou-etal-2024-mystery,cho2025revisitingincontextlearninginference}. Originally studied in text-based LLMs, this paradigm has now been extended to Multimodal LLMs (MLLMs)~\cite{li2025advancingmultimodalincontextlearning,cama,huang2024multimodal}, allowing models to incorporate both visual and textual information in demonstrations. Multimodal ICL has shown promise on various vision-language tasks, such as visual question answering~\cite{cama,huang2024multimodal}, offering more flexible and generalizable multimodal understanding. 

\begin{figure*}[t]
    \centering
    \includegraphics[width=0.95\linewidth]{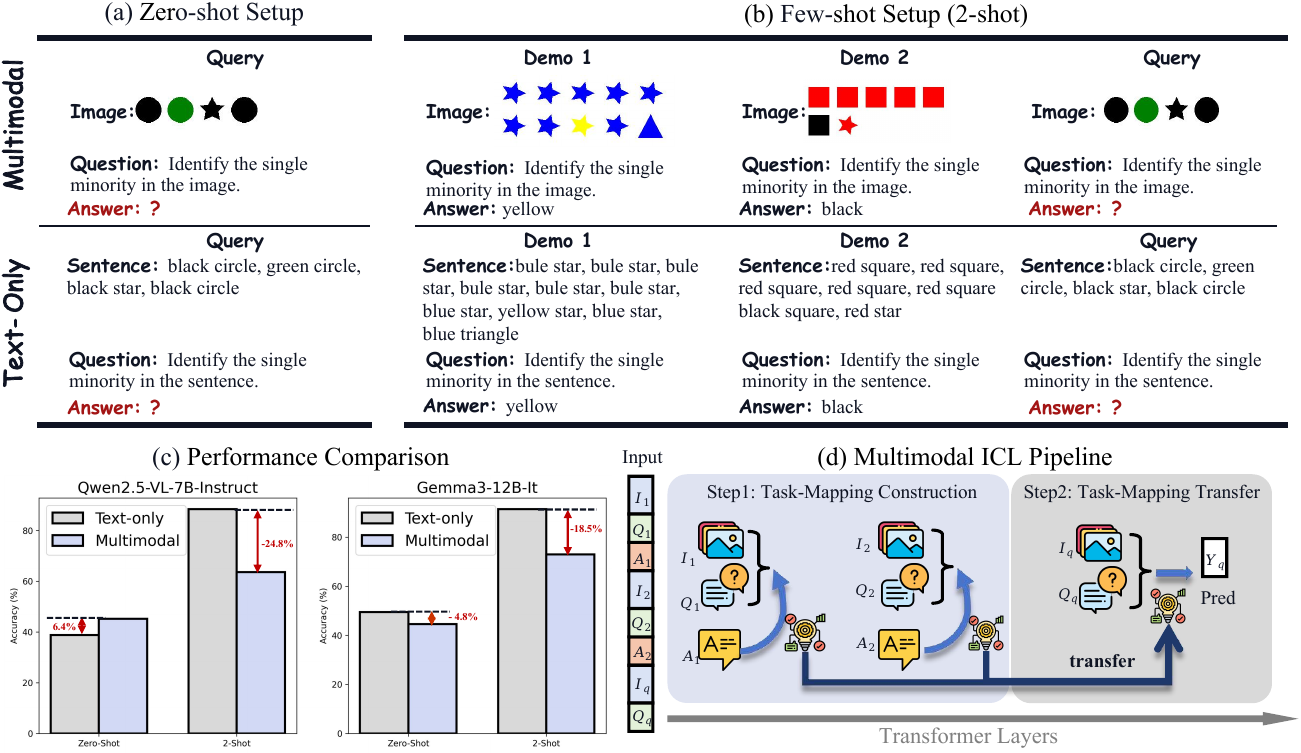}
    \caption{(\textbf{a–b}) Illustration of constructed outlier detection task~\cite{true-micl}, where (a) shows the zero-shot setup and (b) shows the few-shot setup. In the few-shot scenario (2-shot), the model must infer from the demonstrations whether the query should be solved based on shape or color, and then apply this inferred rule to identify the outlier item in either the image or the sentence.
(\textbf{c}) Performance comparison of Qwen2.5-VL-7B and Gemma-3-12B, showing that multimodal ICL suffers a clear degradation under the few-shot setting (4-shot) relative to text-only ICL.
(\textbf{d}) Illustration of the multimodal ICL pipeline, including task mapping construction from demonstrations and task mapping transfer to the query.}
    \label{fig:motivation}
\end{figure*}

Recent studies have investigated multimodal ICL from different perspectives~\cite{qin2024factorsaffectmultimodalincontext,cama,gao2024aimletmultimodallarge,chen2025ocean}, such as the underutilization of visual context~\cite{taco,true-micl} and sensitivity to demonstration configurations~\cite{qin2024factorsaffectmultimodalincontext,chen2024multimodallargelanguagemodels}. In parallel, recent works have introduced token pruning to improve efficiency~\cite{li2025catpcontextuallyadaptivetoken,gao2024aimletmultimodallarge}. Despite these advances, the underlying mechanisms of ICL within the multimodal domain remain poorly explored. It remains an open question \emph{how} MLLMs perform ICL and \emph{whether} this process differs fundamentally from textual ICL. Motivated by these gaps, this paper proposes a systematic analysis framework for multimodal ICL.

First, to investigate the differences between multimodal and textual ICL, we construct an identical ICL dataset (see Fig.~\ref{fig:motivation}(a–b); details in Sec.\ref{sec:diff}). For the same ICL problem, we provide both textual and multimodal formulations~\cite{nikankin2025taskdifferentcircuitsdisentangling}, enabling a controlled comparison of model performance under the two settings (More details see Sec.\ref{sec:diff}). Evaluation on Qwen2.5-VL~\cite{qwen25vl} and Gemma-3~\cite{gemma3} (Fig.~\ref{fig:motivation}(c)) reveals a striking contrast: while zero-shot performance is comparable across modalities, few-shot performance significantly degrades with multimodal demonstrations. This gap indicates that MLLMs struggle to leverage multimodal context as effectively as text.

To delve into why multimodal ICL diverges from its textual counterpart and identify the bottlenecks constraining its performance, we take a closer look at the multimodal ICL pipeline. Specifically, we decompose multimodal ICL into two steps~\cite{taco,cho2025revisitingincontextlearninginference}, as illustrated in Fig.~\ref{fig:motivation}(d): \textit{Task Mapping Construction}, where the model learns and builds the task mapping within the demonstrations, and \textit{Task Mapping Transfer}, where the learned mapping is applied to the query for final prediction. Here, task mapping refers to inference mechanism that transforms questions into their corresponding answers within the MLLMs latent space.

Our analysis yields two pivotal findings. First, we find that MLLMs are capable of constructing task mappings from multimodal demonstrations: in middle layers, demonstration labels are grounded to the correct visual evidence, indicating successful task mapping construction. Notably, this grounding emerges even in cases where final predictions on the query are incorrect, suggesting that task mapping construction alone is insufficient for successful multimodal ICL. Second, we show that these constructed task mappings at mid-layer are not reliably transferred to guide query-time reasoning, due to a {\emph{critical cross-modal  misalignment between visual perception and textual reasoning during ICL}}.
Specifically, attention from the query's last token to the demonstration labels and their grounded visual evidence increases sharply in later layers. However, at these later layers, the task mappings induced by the demonstrations are no longer accurate or reliably represented, preventing the model from applying the inferred task structure during reasoning. Consequently, late-layer predictions are driven primarily by perceptual signals in the query image rather than by the task mappings inferred from demonstrations. In other words, \emph{task mapping construction and task mapping transfer are decoupled across different layers.}

Finally, to access the practical implications of this analysis, we introduce a simple yet effective inference-stage ICL enhancement method. Our approach explicitly extracts the task mapping encoded within the demonstrations and reinforces its transfer to the query by dynamically amplifying attention toward task-relevant evidence while suppressing irrelevant regions. This intervention serves as an exploratory probe that validates the causal role of task mapping transfer in multimodal ICL. In summary, our main contributions are: 
\begin{enumerate}[leftmargin=*, itemsep=2pt]
    \item We present a controlled experimental framework that enables a direct comparison between text-only and multimodal ICL under identical task formulations, revealing a substantial performance gap in few-shot settings.
    \item We conduct a systematic mechanistic analysis of multimodal ICL that decomposes the process into task mapping construction and task mapping transfer, uncovering the mechanisms of multimodal ICL and identifying the bottlenecks that constrain multimodal ICL performance.
    \item We explore a lightweight inference-time intervention that reinforces task mapping transfer, providing empirical evidence for the causal relevance of the identified bottleneck and demonstrating the practical utility of our analysis.
\end{enumerate}

\section{Background and Related Work}
\label{sec:bg}
\textbf{Multimodal In-Context Learning.}
In multimodal ICL, a model $\mathcal{M}$ receives an input sequence consisting of $n$ multimodal demonstrations and a query, and predicts the target label $\hat{Y}_q$, as follows:
\[\hat{Y}_q = \mathcal{M}\big( (I_1, Q_1, Y_1), \dots, (I_n, Q_n, Y_n), (I_q, Q_q) \big),\]
Here, each demonstration is a triplet $(I_i, Q_i, Y_i)$, where $I_i$ is the image, $Q_i$ is the question, and $Y_i$ is the ground-truth label. The query instance consists of an image-question pair $(I_q, Q_q)$.

\paragraph{Related Work on ICL.} While the mechanisms of text-only ICL have been extensively studied~\cite{cho2025revisitingincontextlearninginference,xie2021explanation, dai2023can, wang2023label, han2023explaining, jeoninformation, zheng2024distributed} through pre-training data~\cite{singh2024needs,singh2024transient,han2023understanding}, feature attribution of inputs ~\citep{pan2023context,kossen2024context}, and functional reduction~\citep{zhang2023trained,han2023explaining}, understanding its multimodal counterpart remains an open challenge. For instance, Cho et al.~\cite{cho2025revisitingincontextlearninginference} decompose text-only ICL inference into three primary operations: input text encoding, semantic merging, and feature retrieval and copying.

Recently, research on multimodal ICL has witnessed significant advancements. One line of work primarily investigates demonstration configurations to enhance ICL performance~\cite{ chen2023understanding, li2023configure, yang2024exploring, baldassini2024makes, qin2024factors, xu2024introspection,wu2025empirical,luo2024does,taco,qin2024factorsaffectmultimodalincontext,chen2024multimodallargelanguagemodels,huang2025mimicking}, and establishes comprehensive benchmarks~\cite{zong2024vl,true-micl}. 
Another strand focuses on fine-tuning the MLLMs for better multimodal ICL ability~\citep{zhao2023mmicl,doveh2024towards,li2023otter,li2023mimic,gao2025aim,li2023otter,jia2024symdpo,true-micl}, while a separate line of work investigates inference-stage methods that improve multimodal ICL without additional training~\cite{cama}.
In parallel, some studies focus on improving efficiency by addressing the issue of token redundancy in ICL~~\cite{li2025catpcontextuallyadaptivetoken,gao2024aimletmultimodallarge}.
However, the internal latent mechanisms governing multimodal ICL--and how they fundamentally differ from text-only ICL--remain largely unexplored. Our work fills this gap by conducting the first in-depth analysis to uncover the structural bottlenecks in cross-modal task mapping and transfer.

\section{Multimodal ICL is More Challenging}
\label{sec:diff}
This section provides a systematic comparison between multimodal and text-only ICL under controlled conditions. Inspired by \citet{true-micl,nikankin2025taskdifferentcircuitsdisentangling}, we design a controlled outlier detection task. The objective is to identify a single minority instance that deviates from the majority based on a specific feature (shape or color). The label space consists of 4 shape categories (circle, triangle, square, star) and 10 color categories (yellow, blue, green, red, black, orange, purple, pink, brown, gray). To isolate the effect of modality, we instantiate each problem in this task  in two distinct formats: text-only and multimodal. As illustrated in Fig.~\ref{fig:motivation}, the text-only format presents all task-relevant information in natural language, whereas the multimodal format requires the model to integrate multimodal information, particularly visual evidence, for correct task completion. This design ensures that only the input modality varies while keeping task structure and supervision fixed.

\paragraph{Performance Gap.}  We evaluate representative multimodal large language models under zero-shot and few-shot settings to show that multimodal ICL is fundamentally more difficult than textual ICL. The results in Tab.~\ref{tab:bg_cmp} reveal a clear performance gap under few-shot conditions. For example, Qwen2.5-VL-7B model exhibits an accuracy drop of over \textbf{24}\% when moving from text-only to multi-modal demonstrations. \emph{This degradation indicates that MLLMs struggle to effectively leverage multimodal demonstrations for task induction and transfer.}

\begin{table}[h]
\begin{center}
 \resizebox{0.48\textwidth}{!}{
\setlength{\tabcolsep}{3.5mm}{
\begin{tabular}{lllll}
\toprule
 Model& Size & ICL Type  &  Zero-Shot & 4-shot \\
\hline
\multirow{4}{*}{Qwen2.5-VL} &  \multirow{2}{*}{7B}  & Text Only & 38.80 & \textbf{88.40} \\
& & Multimodal & \textbf{45.20} (\textcolor{blue}{ $\uparrow6.40$}) & 63.60  (\textcolor{blue}{ $\downarrow 24.80$}) \\
\cline{2-5}
&\multirow{2}{*}{32B} & Text Only & \textbf{50.00}& \textbf{80.47}\\
 & &Multimodal &  43.60 (\textcolor{blue}{ $\downarrow 6.40$}) & 76.13 (\textcolor{blue}{ $\downarrow 4.34$}) \\ \hline
\multirow{4}{*}{Gemma-3}&\multirow{2}{*}{12B}& Text Only & \textbf{49.40} & \textbf{91.40} \\
 & & Multimodal &44.60 (\textcolor{blue}{ $\downarrow 4.80$}) & 72.93 (\textcolor{blue}{ $\downarrow 18.47$})\\
 \cline{2-5}
& \multirow{2}{*}{27B}  & Text Only & \textbf{50.00} & \textbf{89.67} \\
  & & Multimodal & 48.00 (\textcolor{blue}{ $\downarrow 2.00$}) &78.80 (\textcolor{blue}{ $\downarrow 10.87$})  \\
\bottomrule
\end{tabular}}}
\end{center}
\caption{Comparison between text-only and multimodal ICL across model families and sizes.}
\label{tab:bg_cmp}
\end{table}


\begin{figure}[h]
    \centering
    \includegraphics[width=0.95\linewidth]{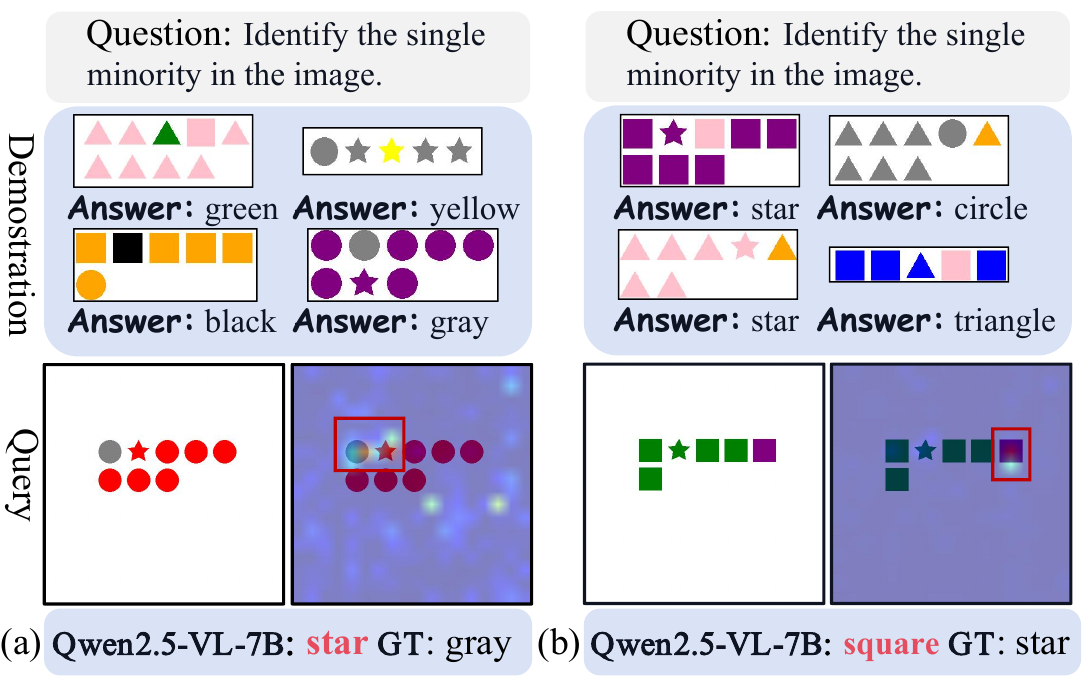}
    \caption{\textbf{Qualitative examples of error cases}. (a) False task recognition: the model misreads the task mapping in the demonstrations and outputs the wrong attribute (shape star) while the true minority is color gray. (b) Correct task recognition but false answer: although the model identifies the correct task mapping (detecting the OOD sample by shape feature), it still predicts the wrong minority because it attends to incorrect regions in the image (square instead of star).}
    \label{fig:case_study}
\end{figure}

\begin{figure}[h]
    \centering
    \includegraphics[width=\linewidth]{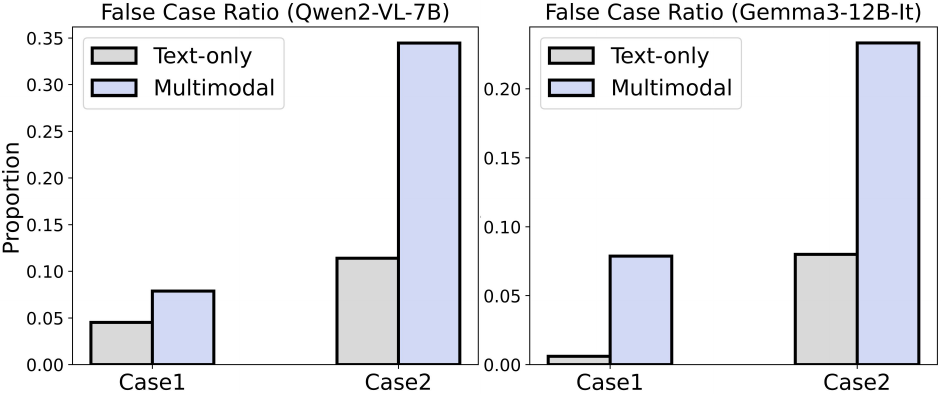}
    \caption{Proportion of error types in text-only ICL and multimodal ICL. Case 1 corresponds to incorrect task recognition, and Case 2 corresponds to correct task recognition but an incorrect answer. }
    \label{fig:false_case}
\end{figure}

\paragraph{Case Study.} We further analyze the error cases of text-only and multimodal ICL. Specifically, we first check whether the model correctly identifies the task type, i.e., whether it performs outlier detection based on the correct feature (color or shape). This is determined by string-matching the model output to see whether the prediction refers to color or shape. We then examine whether the prediction matches the ground-truth label. Based on this analysis, we categorize the errors into two types: (Case 1)~\textit{False Task Recognition} and (Case 2) \textit{Correct Task Recognition but Incorrect Answer}, where Fig.~\ref{fig:case_study} provides representative examples of each error type. Case 1 occurs when the model fails to recognize the intended task, stemming from either a failure to infer the correct mapping from demonstrations or an inability to apply it to the query (e.g., detecting the outlier by shape when the demonstrations imply color). Case 2 occurs when the model correctly infers the task rule but fails to produce a correct final prediction (e.g., predicting \texttt{square} instead of \texttt{star}). 

\paragraph{Understanding the Gap.} We also report the proportions of each error type in Fig.~\ref{fig:false_case}. As shown, multimodal ICL exhibits significantly higher rates for both error types compared to the text-only baseline.
Specifically, these errors suggest distinct internal failure modes: (i) failures to construct a stable task mapping from multimodal demonstrations, and (ii) failures to accurately transfer the recognized task to the query. Understanding these errors is important because effective multimodal ICL requires both capabilities--learning the task structure from demonstrations and reusing this structure during query inference by integrating visual and textual cues into a unified reasoning process. 

However, existing analyses based solely on input-output behavior do not reveal how task mappings are formed \emph{internally} or \emph{why} they fail to generalize from demonstrations to the query~\cite{taco,qin2024factorsaffectmultimodalincontext,chen2024multimodallargelanguagemodels}.
To move beyond performance-level observations, we next examine the internal mechanisms underlying multimodal ICL and identify the structural bottlenecks that hinder performance.

\begin{figure*}[h]
    \centering
    \includegraphics[width=0.9\linewidth]{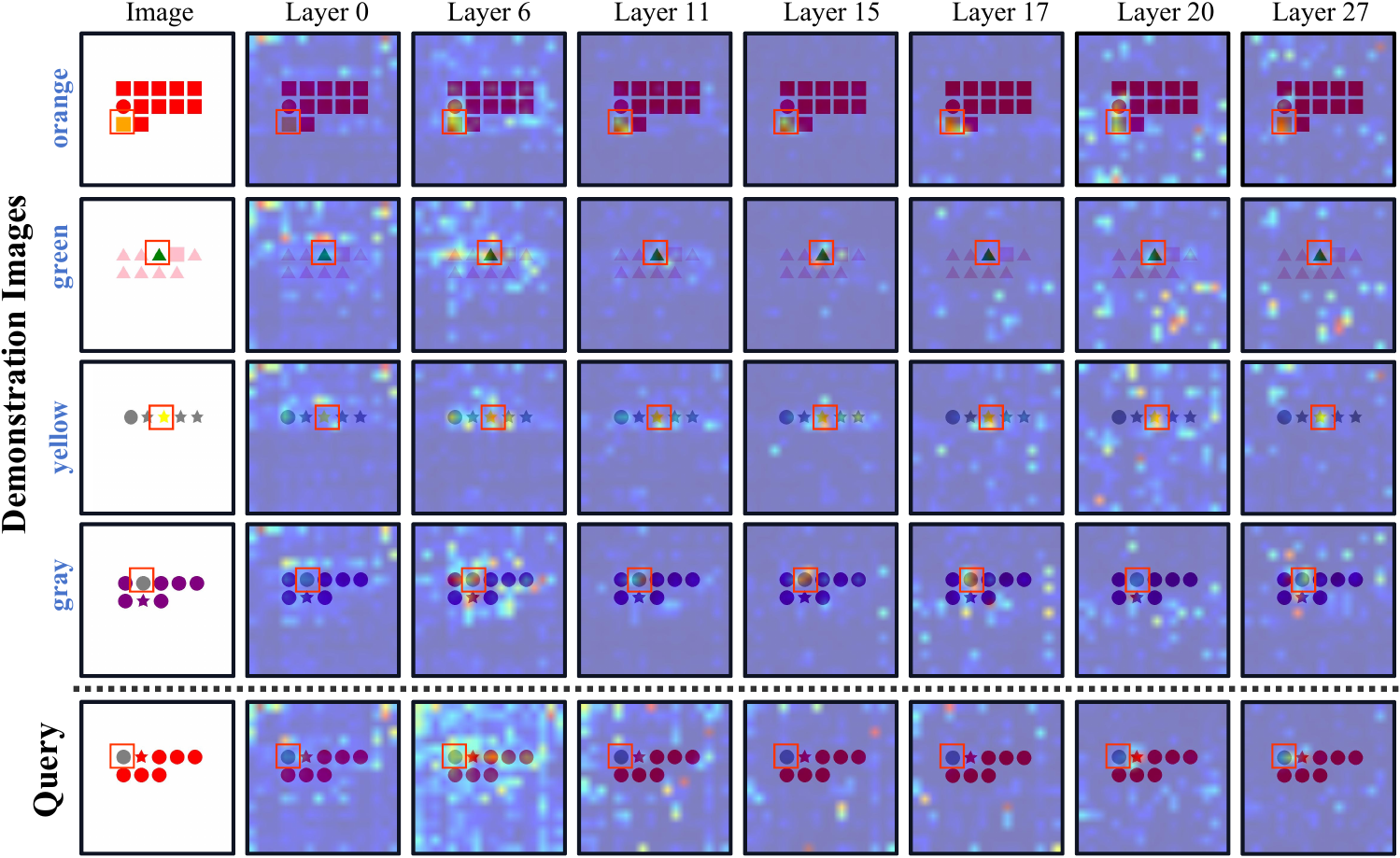}
    \caption{\textbf{Layer-wise visualization of attention from demonstration's labels tokens (or the last token) to image tokens in a multimodal ICL example.} The four demonstrations are labeled by the color outlier, but the model exhibits the ~\textit{False Task Recognition} on the query and incorrectly predicts \texttt{star} instead of \texttt{gray}. \textcolor{red}{Red bounding boxes} denote ground-truth evidence regions. Demonstration label tokens form clear object-level grounding only in mid layers, with early layers being diffuse and deeper layers becoming noisy again. In contrast, the query token shows no meaningful grounding until the final several layers, where localization finally appears.} 
    \label{fig:vis_attn_case}
\end{figure*} 
\section{Understanding Multimodal ICL Mechanisms}
\label{sec:analysis}
To understand the process of multimodal ICL, we conduct a systematic analysis of its internal behaviors. Our goal is to uncover the mechanisms of multimodal ICL and identify the bottlenecks that constrain multimodal ICL performance. Guided by the errors identified in Sec.~\ref{sec:diff}, we organize this section around two core questions:
\textbf{RQ1} evaluates whether MLLMs can establish effective task mappings from demonstrations, and
\textbf{RQ2} studies whether MLLMs can transfer the learned task mappings from demonstrations to the query.

\subsection{RQ1: Can multimodal ICL Establish Effective Task Mappings?}
\label{sec:mapping}
First, we examine whether the failures occur during demonstration-level task mapping construction.
\paragraph{Mid-layer Grounding as Evidence of Task Mapping Construction.} To investigate how task mappings are formed, we begin with a qualitative case study that visualizes attention from demonstration label tokens to image regions across layers (Fig.~\ref{fig:vis_attn_case}). In this example, constructing a task mapping requires grounding each demonstration label to the color attribute of the outlier object. When the attention from a demonstration’s label aligns with the ground truth evidence region in its corresponding image (\textcolor{red}{red bounding boxes}), this indicates that the model correctly grounds the label and successfully builds the underlying task mapping.

As shown in Fig.~\ref{fig:vis_attn_case}, despite the incorrect final prediction, for \textit{demonstrations}, early layers exhibit scattered and unstructured patterns, but the mid layers develop a clear focus on the correct object, indicating the emergence of visual grounding. In deeper layers, this focus becomes less stable and the attention spreads again.

\begin{figure}[t]
    \centering
    \includegraphics[width=\linewidth]{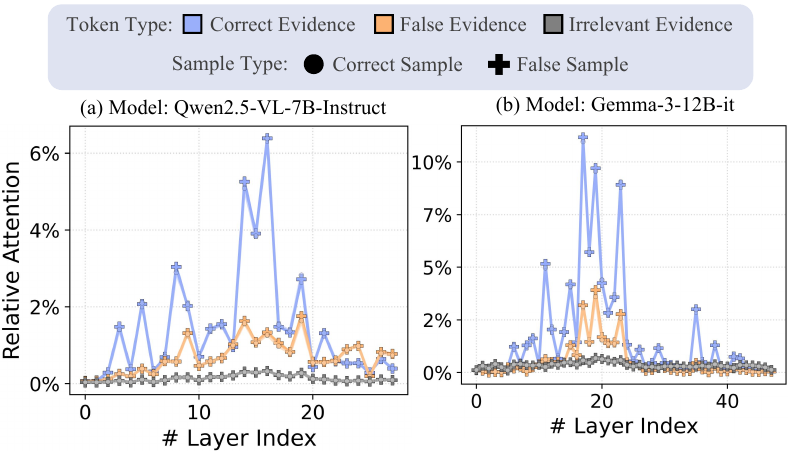}
    \caption{\textbf{Layer-wise attention ratios over different image regions for correct ($\bullet$) vs. incorrect (\textbf{$+$}) predictions.} Relative attention from demonstration label tokens to correct, false, and irrelevant evidence regions. Both models show a strong peak on correct evidence at the midlayer, indicating stable visual grounding within the demonstrations.}
    \label{fig:attn_evidence}
\end{figure}

\paragraph{Quantifying Demonstration-Level Visual Grounding.}
To move beyond individual examples, we quantitatively analyze attention allocation from demonstration label tokens to different image regions across layers, and present the results in Fig.~\ref{fig:attn_evidence}. We divide each image into three evidence types: the correct outlier region as \textit{correct evidence}, the incorrect outlier region as \textit{false evidence}, and all remaining areas as \textit{irrelevant evidence}. For each layer, we compute the relative attention assigned to these regions by the demonstration’s label tokens.

Across models and across both correct and incorrect samples, we observe a pronounced peak of attention to correct evidence regions in intermediate layers, while attention to false and irrelevant regions remains consistently low.
This indicates that the model effectively achieves label-evidence alignment within the demonstrations and are able to construct task mappings in intermediate layers, regardless of whether the final prediction on the query is correct. 

\paragraph{Causal Intervention Study.}
While attention alone does not constitute a complete explanation of model behavior, we complement attention analysis with targeted interventions to establish a causal link. In particular, we conduct an intervention that disrupts visual grounding by replacing attention over image tokens with a uniform distribution. We then compare the model’s multimodal ICL performance before and after the intervention on different model families (Tab.~\ref{tab:attn_inter1}). The results show uniform attention causes a precipitous performance collapse across all model families. Notably, the accuracy drops to levels near or below the zero-shot baseline, demonstrating that the precise attention to visual evidence in intermediate layers is a causal prerequisite for the performance gains observed in multimodal ICL.

\begin{table}[h]
\begin{center}
 \resizebox{0.48\textwidth}{!}{
\setlength{\tabcolsep}{3.5mm}{
\begin{tabular}{l|ll|ll}
\toprule
 Model   &  \multicolumn{2}{c|}{Qwen2.5-VL} &\multicolumn{2}{c}{Gemma-3}   \\
 \hline
Size &7B &  32B  & 12B & 27B \\ 
\hline
Zero-shot &45.20&43.60 &44.60&48.00 \\
Original (4-shot) & 69.09&  78.45 &  72.93& 78.62 \\
\rowcolor{LightGray} After UAS (4-shot) &45.44 & 52.19 & 50.56 & 0.0\\
\bottomrule
\end{tabular}
}}
\end{center}
\caption{\textbf{Causal Intervention Study.} We apply our \textit{Uniform Attention Suppression} (UAS) intervention, which replaces the model’s attention over image tokens with a uniform distribution, leading to significant performance degradation. }
\label{tab:attn_inter1}
\end{table}

\paragraph{Task Mapping Construction Is Necessary but Not Sufficient.} Within the scope of our controlled setup, these findings show that MLLMs are capable of constructing task mappings from multimodal demonstrations by grounding labels to the correct visual evidence in intermediate layers. However, since this grounding occurs even when final predictions fail, demonstration-level task mapping construction alone cannot explain the error patterns observed in Sec.~\ref{sec:diff}. This suggests that the failure of multimodal ICL arises not solely from constructing task mappings, but from transferring or applying these mappings when reasoning over the query—an issue we investigate next.

\begin{takeaway}{Finding 1}
During  multimodal ICL, models can form task mappings within the demonstrations by grounding demonstration labels more to the correct visual evidence in intermediate layers. 

This grounding emerges regardless of whether the final prediction on the query is correct or not.
\end{takeaway}

\subsection{RQ2: Can Task Mappings Be Transferred from Demonstrations to the Query?} 
Successful multimodal ICL requires not only inferring the task structure from demonstrations, but also reusing this structure when reasoning over the query. In this section, we analyze how demonstration-induced task information influences the model’s final decision, and identify the mechanisms that lead to failures in task mapping transfer.

\paragraph{Failure to Access Task Mappings in Intermediate Layers.} To examine whether task mappings are transferred to the query, we measure the attention from the query’s last token to both the demonstration labels and the label-relevant image regions across layers. In text-only ICL, such attention patterns are known to reflect how LLMs incorporate demonstration-induced information when answering the query~\cite{cho2025revisitingincontextlearninginference,xie2021explanation, dai2023can, wang2023label, han2023explaining, jeoninformation, zheng2024distributed}. If multimodal task mappings are successfully transferred, we would expect the query’s final token to attend to the demonstration labels and their grounded visual regions, particularly in layers where task mappings are constructed.

\begin{figure}[t]
    \centering \includegraphics[width=0.8\linewidth]{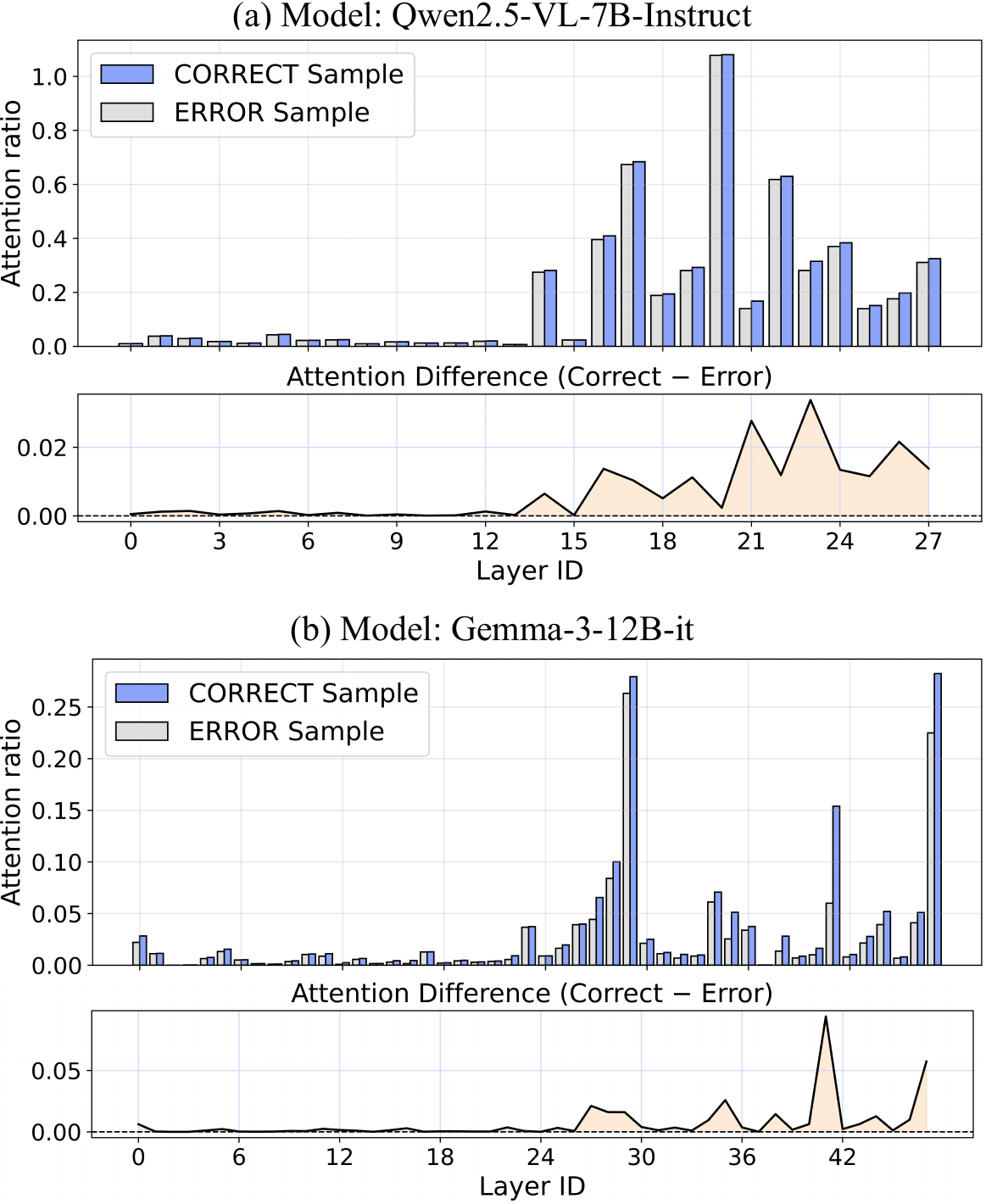}
    \vspace{-10pt}
    \caption{Comparison of Correct and Error Samples in Layer-wise Last-Token Attention to Evidence Regions.}
    \vspace{-10pt}
    \label{fig:attn_query2demo}
\end{figure}

As shown in Fig.~\ref{fig:attn_query2demo}, we find that attention from the last token to demonstration labels and their grounded visual evidence is nearly zero in the middle layers, despite the presence of demonstration-level grounding at these layers (recall Sec.~\ref{sec:mapping}). Meanwhile, it subsequently rises in later layers to drive the final prediction. 
This implies that task information is encoded early but not immediately utilized. In other words, \emph{task mapping construction and task application are decoupled across different layers of the model.}

\paragraph{Perception-Reasoning Misalignment as the Core Bottleneck.} Interestingly, we also observe in Fig.~\ref{fig:attn_query2demo} that attention from query's last token to the demonstration labels and their grounded visual evidence increases sharply in later layers. 
However, at these later layers, the task mappings induced by the demonstrations are no longer accurate or reliably represented. 
As a result, the final prediction is driven primarily by late-layer perceptual cues rather than by demonstration-induced task mappings, causing multimodal demonstrations to provide little meaningful guidance for the decision.
Specifically, we observe that correct samples maintain significantly higher attention to evidence regions in late layers compared to error samples (see the difference plots in Fig.~\ref{fig:attn_query2demo}), implying that strong late-layer perceptual cues are a prerequisite for accurate predictions. This finding also clarifies the error patterns in Sec.~\ref{sec:diff}: when task mappings fail to influence late-layer reasoning, the model may either behave as if the task was not inferred or attend to correct visual evidence without applying the correct decision rule. 

\begin{takeaway}{Finding 2}
Task mappings inferred from multimodal demonstrations fail to reliably influence query-time reasoning. This is primarily due to the \emph{cross-modal misalignment between visual perception and textual reasoning}: {task mappings involving visual contexts are constructed in mid layers, but fail to propagate across the modality gap to the final reasoning process.
In other words, \emph{task mapping construction and task application are decoupled across different layers and the cross-modal barrier.}
} 
\end{takeaway}
\section{Mapping-Guided Inference: Bridging Task Mapping and Transfer}
The analyses in Section~\ref{sec:analysis} reveal a clear mechanistic bottleneck in multimodal in-context learning. In this section, we ask a pragmatic follow-up question: {can insights from this analysis be leveraged to improve multimodal ICL behavior in practice}? 

To this end, we introduce Mapping-Guided Inference (\ours), an inference-time intervention that extracts the latent task mapping encoded within the demonstrations in mid layers and explicitly injects it into the query's attention mechanism. By guiding the query token to attend to evidence regions grounded by demonstration labels in later layers, \ours is designed to bridge the misalignment between
perception and reasoning in multimodal ICL. {We view this approach as an exploratory, analysis-driven intervention that probes the practical utility of the mechanisms identified above, rather than as a fully optimized method.}

\subsection{Estimating Task Mapping from Demonstrations}
As shown in Section~\ref{sec:mapping}, task mappings in multimodal ICL are constructed within demonstrations via mid-layer visual grounding. In \ours, we explicitly extract this latent task mapping from the model’s internal attention patterns.

\paragraph{Constructing the Attention Candidate Set.} We first collect the attention weights from each demonstration's label token to its corresponding all image tokens. For $n$-shot setup at layer $\ell$, we define the attention set $\mathcal{A}^\ell = \big\{\mathbf{a}^{(\ell,h)}_{\mathrm{lbl}_i\to\mathrm{img}_i} \big\}_{i,h}$,  where $\mathbf{a}^{(\ell,h)}_{\mathrm{lbl}_i\to\mathrm{img}_i} \in \mathbb{R}^{L_{\mathrm{img}_i}}$ denotes the attention distribution from the label token of demonstration $i$ to its image tokens at layer $\ell$ and head $h$. Here, $\ell \in \{1, \dots, L\}$ and $h \in \{1, \dots, H\}$ index the Transformer layers and attention heads, respectively, while $L_{\mathrm{img}_i}$ denotes the number of image tokens in the $i$-th demonstration.

\paragraph{Identifying the Peak Grounding Layer.} 
We aim to identify the layer $\ell^*$ that exhibits the strongest visual grounding between demonstration labels and image regions. We quantify grounding strength using the entropy of the attention distribution, where lower entropy indicates a more focused attention over specific visual evidence.

For each layer $\ell$, we compute the average entropy of the attention from demonstration label tokens to their corresponding image tokens, aggregated across all demonstrations and attention heads. The peak grounding layer is defined as:
\vspace{-4pt}
\begin{equation}
    \ell^{*} = \arg\min_{\ell} \sum_{i=1}^{n}\sum_{h=1}^{H}  \mathcal{H}\Big(\mathbf{a}^{(\ell,h)}_{\mathrm{lbl}_i\to\mathrm{img}_i} \Big),
\end{equation}
where $\mathcal{H}(\mathbf{p}) = -\sum_{j} p_j \log p_j$ is the entropy of a probability distribution $\mathbf{p}$. Note that the attention weights $\mathbf{a}$ are normalized by their sum (i.e., $\mathbf{p} = \mathbf{a} / \|\mathbf{a}\|_1$) to form a valid probability distribution over the image tokens. The attention patterns at this peak layer \(\ell^{*}\) are then used as our estimate of the task mapping:
\begin{equation}
     \hat{\mathcal{M}} = \mathcal{A}^{(\ell^{*})}.
     \label{eq:task_mask}
\end{equation}

\subsection{Intervening on Query Attention}
After obtaining the estimated task mapping $\hat{\mathcal{M}}$, we guide the model to answer the query by reusing this mapping. Specifically, when the query's label token attends to the image tokens of demonstration $i$, we encourage it to mimic the attention pattern established by the label of demonstration $i$ at the peak layer.

\paragraph{Injecting the Task Mapping into Query Attention.} We intervene on the attention mechanism in layers deeper than a starting layer $L_{\mathrm{start}}$. For a given layer $\ell > L_{\mathrm{start}}$ and head $h$, we modify the attention from the query token (q) to the image tokens of demonstration $i$ as follows: 
\begin{equation}
\tilde{\mathbf{a}}^{(\ell,h)}_{\mathrm{q}\to\mathrm{img}_i}
    = \mathbf{a}^{(\ell,h)}_{\mathrm{q}\to\mathrm{img}_i}
    + \lambda \cdot \mathbf{a}^{(\ell^{*},h)}_{\mathrm{lbl}_i\to\mathrm{img}_i},
    \label{eq:scale}
\end{equation}
where $\mathbf{a}^{(\ell,h)}_{\mathrm{q}\to\mathrm{img}_i}$ is the original attention from the query to the image tokens of demonstration $i$, $\mathbf{a}^{(\ell^{*},h)}_{\mathrm{lbl}_i\to\mathrm{img}_i} \in \hat{\mathcal{M}}$ denotes the label-to-image attention extracted at the peak grounding layer $l^{*}$, and $\lambda > 0$ controls the strength of the intervention.

After the intervention, we re-normalize the attention scores $\hat{\mathbf{a}}$ to ensure they form a valid probability distribution: $\tilde{\mathbf{a}} = \hat{\mathbf{a}} / \sum \hat{\mathbf{a}}$. This process explicitly encourages the query token to ``look at'' the demonstration images in the same way that defined the task in the demonstrations, thereby promoting consistent visual grounding between demonstrations and query inference.

\subsection{Experiments}

\begin{table}[t]
\centering
\setlength{\tabcolsep}{2pt}
\resizebox{\linewidth}{!}{
\begin{tabular}{l|c|l|llll}
\toprule
 & Size & Method & \textbf{Outlier} & \textbf{Clock} & \textbf{Operator} &  \textbf{OK-VQA}\\
\midrule
\multirow{4}{*}{\cellcolor{white}\centering\rotatebox{90}{Qwen2.5-VL}} &\multirow{2}{*}{7B} & Vanilla &69.09\textcolor{gray}{$_{\pm0.81}$} & 63.45\textcolor{gray}{$_{\pm0.00}$} &77.59\textcolor{gray}{$_{\pm0.00}$} & 48.13\textcolor{gray}{$_{\pm0.11}$} \\
& \multirow{-2}{*}{\cellcolor{white}7B} &\cellcolor{LightGray}  Ours &\cellcolor{LightGray}\textbf{70.17}\textcolor{gray}{$_{\pm0.00}$} & \cellcolor{LightGray}\textbf{64.49}\textcolor{gray}{$_{\pm0.00}$} &\cellcolor{LightGray}\textbf{77.93}\textcolor{gray}{$_{\pm0.00}$}  & \cellcolor{LightGray}\textbf{48.17}\textcolor{gray}{$_{\pm0.27}$}\\
\cmidrule{2-7}
 & 
\multirow{2}{*}{32B} & 
Vanilla & \textbf{78.45}\textcolor{gray}{$_{\pm0.00}$} &74.14\textcolor{gray}{$_{\pm0.00}$}& 97.93\textcolor{gray}{$_{\pm0.00}$}  & 52.67\textcolor{gray}{$_{\pm0.64}$} \\
 & \multirow{-2}{*}{\cellcolor{white}32B} & \cellcolor{LightGray}Ours & \cellcolor{LightGray}\textbf{78.45}\textcolor{gray}{$_{\pm0.00}$} & \cellcolor{LightGray}\textbf{74.48}\textcolor{gray}{$_{\pm0.00}$} &\cellcolor{LightGray} \textbf{98.28}\textcolor{gray}{$_{\pm0.00}$}  & \cellcolor{LightGray}\textbf{53.13}\textcolor{gray}{$_{\pm0.10}$} \\
\midrule
 &\multirow{2}{*}{12B} &
Vanilla & 72.93\textcolor{gray}{$_{\pm0.00}$} & \textbf{66.90}\textcolor{gray}{$_{\pm0.00}$} & 47.93\textcolor{gray}{$_{\pm0.00}$} & 48.39\textcolor{gray}{$_{\pm0.75}$} \\
\rowcolor{LightGray}
\cellcolor{white}&
\multirow{-2}{*}{\cellcolor{white}12B} &
Ours & \textbf{73.85}\textcolor{gray}{$_{\pm0.14}$} & \textbf{66.90}\textcolor{gray}{$_{\pm0.00}$} & \textbf{48.28}\textcolor{gray}{$_{\pm0.00}$} & \textbf{49.51}\textcolor{gray}{$_{\pm0.26}$}  \\
\cmidrule{2-7}
 &
\multirow{2}{*}{27B} &
Vanilla & 78.62\textcolor{gray}{$_{\pm0.00}$} & 76.55\textcolor{gray}{$_{\pm0.00}$} & 70.00\textcolor{gray}{$_{\pm0.00}$}& 48.37\textcolor{gray}{$_{\pm0.63}$}  \\
\rowcolor{LightGray}
\multirow{-4}{*}{\cellcolor{white}\rotatebox{90}{Gemma3}} &
\multirow{-2}{*}{\cellcolor{white}27B} &
Ours & \textbf{79.54}\textcolor{gray}{$_{\pm0.16}$}  & \textbf{76.90}\textcolor{gray}{$_{\pm0.00}$} & \textbf{70.34}\textcolor{gray}{$_{\pm0.00}$}& \textbf{49.29}\textcolor{gray}{$_{\pm0.17}$} \\
\bottomrule
\end{tabular}}
\vspace{-10pt}
\caption{Results of 4-shot multimodal ICL. Experiments are averaged over 3 different seeds, and the performance is in percentage with the standard deviations. Rows shaded in gray correspond to MGI (ours).}
\label{tab:main}
\vspace{-5pt}
\end{table}

\paragraph{Dataset and Models.} We evaluate \ours on the TrueMICL benchmark~\cite{true-micl} (Outlier Detection, Clock Math, Operator Induction) and a natural-image datasets (OK-VQA~\cite{okvqa}). For evaluation, we use exact matching for numerical answers and keyword matching for text. Experiments are conducted on Qwen2.5-VL~\cite{qwen25vl} and Gemma3~\cite{gemma3} across different model scales to ensure robustness. More details refer to Appendix~\ref{sec:appendix_data}.

\paragraph{Main Results.} As shown in Tab.~\ref{tab:main}, \ours consistently enhances multimodal in-context learning performance across various architectures and datasets. The method is particularly effective on tasks requiring strict visual reasoning, contributing to gains in OK-VQA  and Outlier. We present detailed hyperparameter analysis in Appendix~\ref{app:hyper}. Overall, these findings highlight that bridging the gap between mid-layer mapping construction and query inference serves as a promising enhancement.

\section{Conclusion}
In this work, we presented a systematic mechanistic analysis of multimodal ICL. By decomposing multimodal ICL into task mapping construction and task mapping transfer, we identified a core bottleneck: while current multimodal models can construct task mappings by grounding labels to relevant visual evidence, these mappings are not reliably transferred to guide query-time reasoning, due to a misalignment between perception and reasoning. Meanwhile, we introduced a simple inference-time intervention that reinforces task mapping transfer. This lightweight intervention provides empirical evidence for the causal role of task mapping transfer and yields consistent improvements in multimodal ICL performance. More broadly, our findings suggest that effective multimodal ICL requires mechanisms that preserve task structure across layers, and point to future directions in model design and training aimed at better aligning perception and reasoning in MLLM.

\newpage
\section*{Acknowledgments}
We thank Changdae Oh, Seongheon Park, Samuel Yeh for their valuable comments on the manuscript.
This work is supported in part by the AFOSR Young Investigator Program under award number FA9550-23-1-0184, National Science Foundation under awards IIS-2237037 and IIS-2331669, Office of Naval Research under grant number N00014-23-1-2643, Schmidt Sciences Foundation, Open Philanthropy, Alfred P. Sloan Fellowship, and gifts from Google and Amazon.

\section*{Ethical Considerations}
This work focuses on analyzing the internal mechanisms of multimodal large language models in in-context learning settings. Our study is empirical and diagnostic in nature, aiming to improve the understanding of cross-modal task mapping rather than training new foundation models. All datasets used in our experiments are publicly available benchmarks for visual question answering and do not contain personally identifiable or sensitive information. In addition, we do not collect new data or involve human subjects in this work.

\section*{Limitation}
Despite these improvements, our work points to important avenues for future research. First, as an inference-time intervention, \ours operates by guiding the model to utilize its existing mid-layer representations; however, it cannot fundamentally rectify the inherent architectural misalignment between perception and reasoning modules in MLLMs. We hope that our analysis and findings provide a solid foundation for understanding multimodal ICL dynamics and inspire more effective approaches for multimodal adaptation. Second, the method introduces additional hyperparameters (e.g., intervention strength) that may require tuning. To facilitate practical application, we provide comprehensive ablation studies and recommended settings in the Appendix~\ref{app:hyper}.

\bibliography{acl_latex}

\appendix
\label{sec:appendix}
\section{Dataset and Evaluation Details}
\label{sec:appendix_data}

\subsection{Controlled Outlier Detection Dataset.}
To provide a systematic comparison between multimodal and text-only ICL under controlled conditions, we construct a controlled outlier detection task~\citep{true-micl,nikankin2025taskdifferentcircuitsdisentangling}. The core objective is to identify a single minority instance—the outlier—that differs from the majority group based on a designated attribute, specifically \textit{shape} or \textit{color}, as illustrated in Fig.~\ref{fig:motivation} (a-b). The label space is defined by 4 distinct shape categories (circle, triangle, square, star) and 10 color categories (e.g., yellow, blue, red, etc.). To isolate the effect of modality, we instantiate each problem in this task in two distinct formats: text-only and multimodal, as illustrated in Fig.~\ref{fig:motivation} (a-b).

\paragraph{Dataset Statistics.} In total, the dataset contains 2,000 samples, consisting of 1,000 color-based and 1,000 shape-based instances. For evaluation purposes, we randomly select 500 samples per category to form the query set. The remaining samples constitute the support pool. In our experiments, we adopt an $n$-shot setting where each query is paired with $n$ demonstrations sampled from the corresponding support pool to facilitate in-context learning.
\begin{figure*}[h]
    \centering
    \includegraphics[width=\linewidth]{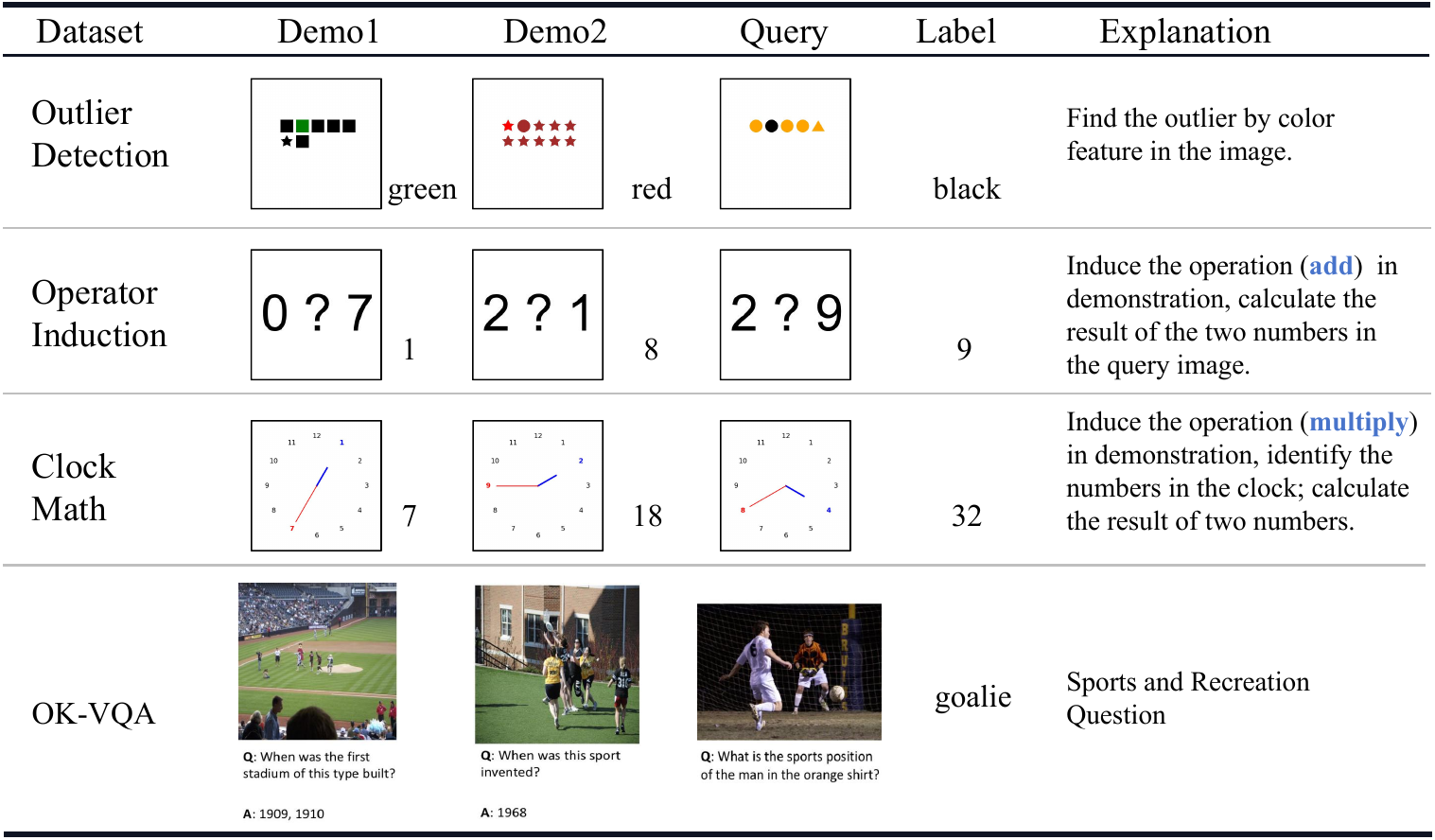}
    \caption{\textbf{An overview of datasets in our papers.} For the datasets sourced from TrueMICL~\cite{true-micl} (Outlier Detection, Operator Induction, Clock Math), the label to the query requires the model to learn the relationship between images and text in the demos. Meanwhile, for the natural-image datasets (OK-VQA~\cite{okvqa}), the model can leverage query-relevant demonstrations to enhance inference on the query instance.}
    \label{fig:dataset}
\end{figure*}

\subsection{Multimodal In-Context Learning Dataset} To validate the effectiveness of \ours, we evaluate it on the TrueMICL benchmark~\cite{true-micl} (comprising Outlier Detection, Clock Math, and Operator Induction) and a natural-image datasets (OK-VQA~\cite{okvqa}). An overview of all datasets used in this work is presented in Fig.~\ref{fig:dataset}:

\begin{itemize}
\item \textit{Outlier Detection} is designed to assess the model's variable binding capability. Each image depicts varying quantities of objects with two distinct shapes and colors. The model is required to identify the target outlier attribute based on the textual demonstration. The label space is defined by 4 distinct shape categories (circle, triangle, square, star) and 10 color categories (yellow, blue, green, red, black, orange, purple, pink, brown, gray).

\item \textit{Operator Induction} consists of simple arithmetic equations formed using three basic operators: addition, subtraction, and multiplication. As the specific operator for each image is not explicitly stated, the model must infer the underlying arithmetic rule by analyzing the relationship between the image content and the target answer.

\item \textit{Clock Math} is a task inspired by arithmetic pattern recognition. To correctly answer the query, the model must first decipher the time displayed on the clock face and subsequently deduce the implicit relationship (e.g., addition) between the time and the answer.

\item \textit{OK-VQA} includes 9,055 training samples and 5,000 validation samples. This dataset challenges models to integrate external knowledge beyond the immediate image and context to generate accurate answers.

\end{itemize}

\paragraph{Dataset Statistics.} Given the limited number of samples per task in the TrueMICL benchmark~\cite{true-micl}, we adopted its format to generate new questions and images, thereby enlarging the dataset scale. Consistent with TrueMICL~\cite{true-micl}, we partitioned the samples in each dataset into a support set of 50 samples and a test set of 290 queries. For the natural-image datasets (OK-VQA~\cite{okvqa}), we randomly sampled 2,048 instances from the validation set to serve as the test set. Furthermore, following standard multimodal ICL evaluation protocols, we utilized Retrieval-based In-Context Example Selection (RICES) to retrieve relevant instances from the training dataset as demonstrations for each query. 

\subsection{Evaluation Metrics} For evaluation, we use exact matching for numerical answers and keyword matching for text. For OK-VQA~\cite{okvqa}, we report VQA accuracy,
i.e., exact match accuracy over a set of ground truth answers. Experiments are conducted on Qwen2.5-VL~\cite{qwen25vl} and Gemma3~\cite{gemma3} across different model scales to ensure robustness.  \\

\section{Models and Implementation Details}
\paragraph{MLLMs.} In this paper, we evaluate the effectiveness and generality of our \ours using recent models from two representative VLM families: Qwen2.5-VL~\cite{qwen25vl} and Gemma 3~\cite{gemma3}. For reproducibility, we list all models with their publicly available checkpoints on Hugging Face\footnote{\textbf{Qwen2.5 VL 7B}https://huggingface.co/Qwen/Qwen2.5-VL-7B-Instruct \\ https://huggingface.co/Qwen/Qwen2.5-VL-32B-Instruct\\ https://huggingface.co/google/gemma-3-12b-it \\ https://huggingface.co/google/gemma-3-27b-it }. We implement all experiments using the \texttt{transformers} and \texttt{PyTorch} libraries on two NVIDIA A100 (80GB) GPUs under \texttt{bfloat16 mixed precision}. For our attention analysis, we use the setting ~\texttt{attn\_implementation='eager'} to retrieve raw attention outputs.

We also attempted to include LLaVA-Next~\cite{liu2024llavanext} and InternVL 3.5 ~\cite{chen2024internvl}; however, these models exhibited significant memory leakage when extracting raw attention maps during inference. Specifically, even the 7B/8B parameter variants caused out-of-memory (OOM) errors on an 80GB A100 GPU under bfloat16 precision. Consequently, we exclude them from the current analysis and leave the resolution of these implementation constraints to future work.

\paragraph{Implement Details of Our \ours.} 
In the main text, we formulated the attention intervention as an additive process (Eq.~\ref{eq:scale}) to provide an intuitive understanding of the mechanism. However, empirically, we observed that directly adding the attention maps yielded suboptimal performance. In our actual implementation, we adopt a selective scaling strategy. Guided by the estimated task mapping $\hat{\mathcal{M}}$ in Eq.~\ref{eq:task_mask}, we identify the specific image tokens that serve as visual evidence for the demonstration label—the critical features the model should attend to when answering the query.

Specifically, we first calculate the mean attention value $\mu$ of the label-to-image attention map $\mathbf{a}^{(\ell^{*},h)}_{\mathrm{lbl}_i\to\mathrm{img}_i}$. We then generate a set of salient indices $\mathcal{S}$ containing only the tokens with attention scores exceeding $k$ times this average:
\begin{equation}
\begin{aligned}
 \mathcal{S} = \left\{ j \mid \mathbf{a}^{(\ell^{*},h)}_{\mathrm{lbl}_i\to\mathrm{img}_i}[j] > k \cdot \mu \right\}, \\ \quad \text{where } \mu = \frac{1}{N} \sum_{j=1}^{N} \mathbf{a}^{(\ell^{*},h)}_{\mathrm{lbl}_i\to\mathrm{img}_i}[j] 
\end{aligned}
\end{equation}
where $N$ is the number of image tokens. Finally, we amplify the query's attention towards these evidence tokens by a factor $\lambda$ while leaving others unchanged:
\begin{equation}
\tilde{\mathbf{a}}^{(\ell,h)}_{\mathrm{q}\to\mathrm{img}_i}[j] = 
\begin{cases} 
\lambda \cdot \mathbf{a}^{(\ell,h)}_{\mathrm{q}\to\mathrm{img}_i}[j] & \text{if } j \in \mathcal{S} \\
\mathbf{a}^{(\ell,h)}_{\mathrm{q}\to\mathrm{img}_i}[j] & \text{otherwise},
\end{cases}
\end{equation}
Where the strength of the intervention $\lambda>1$. 

After this intervention, we re-normalize $\tilde{\mathbf{a}}$ to ensure it forms a valid probability distribution. In our experiments, we typically set $k=1.5$.

\section{Additional Analysis Study}
\subsection{Generality beyond controlled synthetic settings.}
\paragraph{Setup.}
To evaluate whether our observations extend beyond controlled synthetic settings, we conduct additional experiments on a more realistic multimodal reasoning benchmark derived from MM-Vet. Specifically, we sample 50 VQA instances and manually annotate ground-truth evidence regions for each question-answer pair. We analyze layer-wise attention patterns on both Qwen2.5-VL (7B/32B)~\cite{qwen25vl} and Gemma-3 (12B/27B)~\cite{gemma3}. For each layer, we measure (i) attention from demonstration
label tokens to ground truth evidence regions versus other regions, and (ii) attention from the query’s final token to demonstration labels and their corresponding visual evidence (refer to Sec.~\ref{sec:mapping}).

\paragraph{Mid-layer construction of grounded mappings.}
We first examine how attention to ground truth evidence regions evolves across layers. Across all models, we observe a consistent stage-wise pattern. Early layers exhibit uniformly low and comparable attention to both correct and irrelevant regions. In contrast, middle layers show a sharp concentration on grounded regions, where attention to correct regions becomes significantly higher (typically 3--9$\times$) than to other regions. 

However, this separation diminishes again in later layers, where attention to grounded regions drops and becomes comparable to or even lower than that to irrelevant regions. For example, in Gemma-3-12B, attention to correct regions peaks in middle layers (e.g., L11--L23: $\sim$0.02--0.05 vs.\ $\sim$0.003--0.006), but declines in later layers (e.g., L30+: $\sim$0.0015 vs.\ $\sim$0.0032). 

This suggests that multimodal task mappings are strongly constructed in mid-layers but are not reliably preserved during the later stages of inference.

\paragraph{Late-layer consumption and temporal mismatch.}
We further analyze how the query’s final token attends to demonstration labels and their associated visual evidence. If task mappings were successfully transferred, the query token would attend to these elements within the same layers where mappings are formed.

Instead, we observe a temporal mismatch. In early and middle layers, attention from the query token to demonstrated evidence remains weak and diffuse. This attention increases sharply only in much later layers, where the query begins to strongly attend to labels and grounded regions. 

For instance, in Gemma-3-12B, attention remains low in middle layers (e.g., L10--L20: $\sim10^{-4}$), but rises significantly in later layers (e.g., L29+: $\sim10^{-3}$). Notably, this surge occurs after the layers identified as responsible for constructing task mappings.

\paragraph{Summary.}
These results consistently reveal a \emph{mid-layer construction vs.\ late-layer transfer failure} pattern across all evaluated architectures. While multimodal models successfully construct grounded task mappings in middle layers, these mappings are not effectively preserved or utilized during the final stages of inference. This phenomenon generalizes beyond synthetic settings and holds across both Qwen2.5-VL and Gemma-3 models.

\subsection{Attention-Based Analysis of Text and Visual Token Contributions}
\begin{figure}[h]
    \centering
    \includegraphics[width=\linewidth]{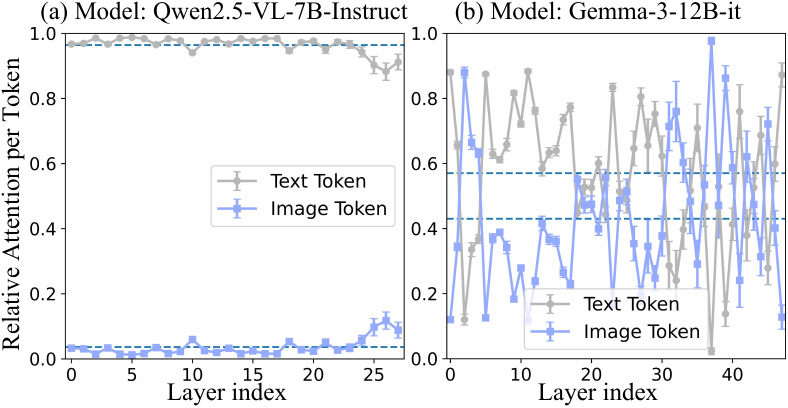}
    \caption{\textbf{Layer-wise Relative Attention per Token of demonstration text and image tokens for different MLLMs families under the 4-shot setup.} Qwen2.5-VL-7B shows a consistently text-dominant pattern across all layers, while Gemma-3-12B displays a modality-switching pattern.} 
    \label{fig:rapt}
\end{figure}

\noindent\textbf{Setup.} Recent work has shown that MLLMs~\cite{true-micl,li2025catpcontextuallyadaptivetoken}, on average, assign significantly less attention to visual tokens compared to textual ones in demonstrations. Here, we take a step further by examining how this imbalance evolves across layers. For each layer, we compute the Relative Attention per Token~\cite{liu2025seeingbelievingprobingdisconnect} (RAT), defined as the ratio between the average attention from the last token (query) to each image or text token in the demonstrations (key). This metric reflects how much attention each modality receives on a per-token basis, rather than describing the overall distribution of attention. Here, we compute the RAT on Qwen2.5-VL-7B and Gemma-3-12B under the 4-shot setup, and present the results in Fig.~\ref{fig:rapt}. 

\noindent\textbf{Layer-wise Attention Allocation to Image and Text Demonstrations.} The results in Fig.~\ref{fig:rapt} reveal clear imbalances in how the model allocates attention to different modalities within the demonstrations. Across models, we observe two distinct patterns in how attention is allocated to text versus image demonstration tokens: (a) \textit{Text-Dominant Pattern.} Qwen2.5-VL-7B exhibits a stable text-dominant pattern across all layers, with visual tokens receiving only negligible attention; (b) \textit{Modality-Switching Pattern.} Gemma-3-12B displays a modality-switching pattern, where attention alternates between text-dominated and image-dominated layers rather than integrating both modalities jointly. Despite their differences, both behaviors reflect the same underlying limitation: the model does not integrate textual and visual information from the demonstrations in a balanced or parallel manner. Instead, attention is assigned to one modality at a time, resulting in sequential rather than fused Multimodal processing.
\begin{figure}[h]
    \centering
    \includegraphics[width=\linewidth]{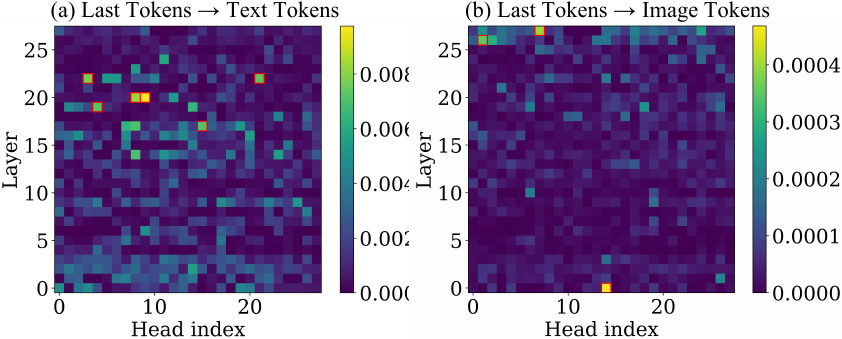}
    \caption{\textbf{Layer-wise Attention Head Activation Patterns on Image and Text Demonstrations under the 4-shot setup.}  Only a few heads show noticeable attention, while most heads remain inactive.}
    \label{fig:attn_head}
\end{figure}

\paragraph{Layer-wise Attention Head Behavior on Demonstrations.} To further examine how the model uses the demonstrations, we analyze how each attention head behaves under the 4-shot setup (see Fig.~\ref{fig:attn_head}). By inspecting a randomly selected example, we track the attention from the last token to the demonstration text and image tokens across all layers and heads. The heatmaps show that only a few heads exhibit meaningful attention to demonstration text, while attention to demonstration image tokens is even sparser. Most heads remain nearly inactive for both modalities. \textbf{\emph{These results indicate that multimodal ICL relies on only a small subset of specialized heads rather than broad distributed attention.}}

\subsection{Hyper-Parameter Analysis}
\label{app:hyper}
To ensure the robustness and optimal performance of \ours, we conduct a comprehensive sensitivity analysis on its two key hyperparameters: the start layer $L_{start}$ and the intervention strength $\lambda$. All experiments in this section are performed using the Qwen2.5-VL-7B model on the Outlet Detection dataset.
\begin{table}[h]
\centering
\setlength{\tabcolsep}{2pt}
\resizebox{\linewidth}{!}{
\begin{tabular}{l|ccccccccc}
\toprule
 $\lambda$& 1 & 1.5 & 2 & 2.5 & 3& 3.5  & 4 &4.5 & 5\\
\midrule 
 Acc. & 68.28 & 68.28 & \textbf{68.97} & 68.28 & 68.28 & 68.62 & 68.62 &\textbf{68.97} & 67.93\\
\bottomrule
\end{tabular}}
\vspace{-10pt}
\caption{Effect of the strength of the intervention $\lambda$.} 
\label{tab:lambda}
\end{table}

\paragraph{Effect of the strength of the intervention $\lambda$.} 
We first analyze the impact of the steering vector's magnitude ($\lambda > 0$). We vary $\lambda$ within the range of $\{1.5, \dots, 5\}$ while we fix $L_{start}$ to a constant value (e.g., $L_{start}=14$). As shown in Table~\ref{tab:lambda}, we observe a trade-off: a small $\lambda$ may not provide enough guidance to correct the model's behavior, whereas an excessively large $\lambda$ risks dominating the original features and degrading generation quality. The empirical results indicate that $\lambda = \text{2}$ yields the best balance between effectiveness and stability.
Based on these empirical results, we adopt $\lambda = 2$ for all datasets in the TrueMICL benchmark. However, for the two VQA datasets, we find that a stronger intervention is necessary to effectively steer the model, thus, we set $\lambda=6$ for those tasks.

\begin{table}[h]
\centering
\setlength{\tabcolsep}{2pt}
\resizebox{\linewidth}{!}{
\begin{tabular}{l|ccccccccc}
\toprule
 $\lambda$& 11 & 12 & 13 & 14 & 15 & 16  & 17 & 18 & 19\\
\midrule 
 Acc. & 68.28 & 67.93 & 68.02 & \textbf{68.97} & 68.28 & 68.28 & 67.93 & 67.93 & 67.93\\
\bottomrule
\end{tabular}}
\vspace{-10pt}
\caption{Effect of the start layer $L_{start}$.}
\label{tab:start_layer}
\end{table}

\paragraph{Effect of the start layer $L_{start}$.} 
Next, we evaluate the influence of the layer where the intervention begins. Guided by the hypothesis that intermediate layers play a crucial role in forming the target concept, we focus our search on the middle block of the model, specifically exploring $L_{start} \in \{11, 12, \dots, 19\}$. We keep the strength of the intervention fixed at the optimal value identified previously, \emph{i.e.} $\lambda=2$. The results, summarized in Table~\ref{tab:start_layer}, demonstrate that the performance is sensitive to the insertion depth, with the optimal performance achieved at layer $L_{start} = \text{14}$. This suggests that intervening too early may disrupt low-level feature extraction, while intervening too late may fail to sufficiently steer the model's high-level reasoning. Meanwhile, the optimal insertion depth varies depending on the total depth of the model architecture. Specifically, for the \textit{Qwen2.5-VL} family, we set $L_{start}=16$ for the 7B model and $L_{start}=40$ for the 32B model. Similarly, for the \textit{Gemma-3} family, we utilize $L_{start}=17$ for the 12B model and $L_{start}=37$ for the 27B model. 

\begin{figure*}[t]
    \centering
    \includegraphics[width=0.9\linewidth]{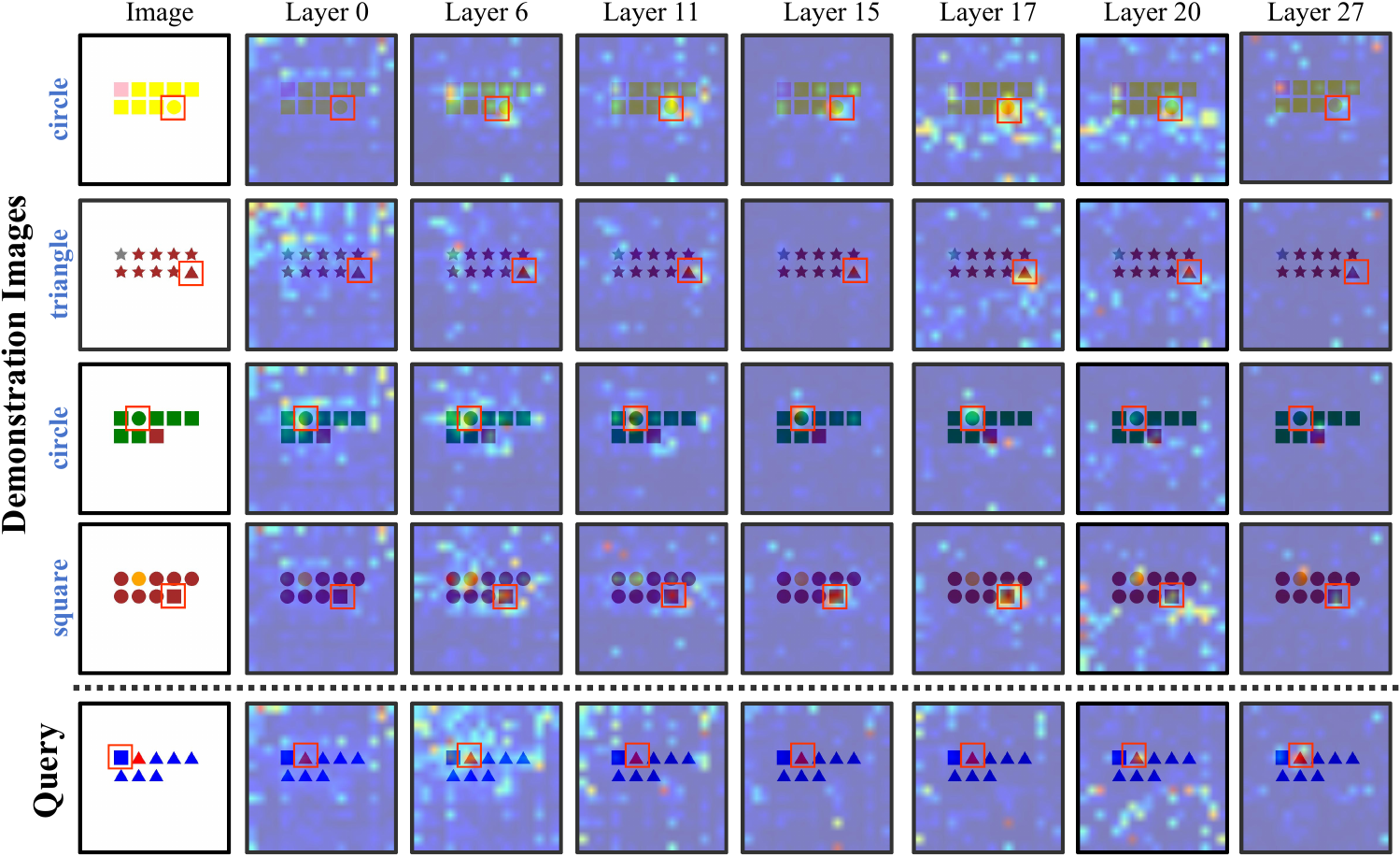}
    \caption{\textbf{Additional layer-wise visualization of attention from demonstration label tokens (or the last token) to image tokens in a multimodal ICL example.} The four demonstrations are labeled by the shape outlier, yet the model exhibits \textit{Correct Task Recognition but False Answer} on the query, incorrectly predicting \texttt{triangle} instead of \texttt{square}. Red bounding boxes denote \textcolor{red}{ground-truth evidence regions}. Demonstration label tokens form clear object-level grounding only in mid layers, with early layers being diffuse and deeper layers becoming noisy again. In contrast, the query token shows no meaningful grounding until the final several layers, where localization finally appears.} 
    \label{fig:more_vis_attn_case}
\end{figure*}

\paragraph{Effect of varying the number of demonstrations.}
We conduct experiments to analyze how the number of demonstrations ($k$) affects model performance. Specifically, we compare our MGI method with a baseline on the Outlier Detection dataset using Qwen2.5-VL-7B, where $k\in [1,2,3,4]$.

\begin{table}[t]
\centering
\small
\begin{tabular}{c|cccc}
\toprule
\textbf{Setup} & \textbf{1-shot} & \textbf{2-shot} & \textbf{3-shot} & \textbf{4-shot} \\
\midrule
Baseline & 65.52\% & 66.55\% & 67.59\% & 69.09\% \\
Ours (MGI) & 66.90\% & 67.59\% & 67.93\% & 70.17\% \\
\bottomrule
\end{tabular}
\caption{Effect of the number of demonstrations ($k$) on Outlier Detection performance with Qwen2.5-VL-7B.}
\label{tab:demo_scaling}
\end{table}

Notably, while performance improves as the number of demonstrations increases, the gains remain relatively modest, indicating that vanilla multimodal in-context learning (MM-ICL) struggles to fully utilize additional examples. In contrast, MGI consistently enhances demonstration utilization, yielding stable improvements across all values of $k$.

\subsection{Inference Time Cost Analysis}
We evaluated the inference latency of \ours using Qwen2.5-VL-7B on \textit{Outlier Detection} dataset. The results indicate that our method maintains high efficiency during the generation phase, with an average inference time of \textbf{842.41 ms}, which is comparable to the baseline's \textbf{800.11 ms}. This negligible overhead ($\sim$5\%) demonstrates that the proposed attention intervention (Eq.~\ref{eq:scale}) is lightweight and does not hinder the token decoding speed.

However, the primary increase in total latency stems from the preparation phase, specifically identifying the peak grounding layer for each query, which incurs a setup cost of \textbf{876.34 ms}. This step involves aggregating full attention distributions to compute entropy. Crucially, this is a \textbf{per-query pre-computation} that depends solely on the input context and does not accumulate with the number of generated tokens, ensuring the method remains efficient even when generating long responses.

\subsection{Additional Layer-wise Visualization of Attention}
\label{sec:appendix_attn}
In this section, we present an additional case study visualizing the attention from demonstration label tokens to image regions across layers. In contrast to the \textit{False Task Recognition} case discussed in the main text, Figure~\ref{fig:vis_attn_case} illustrates a different failure mode: \textit{Correct Task Recognition but False Answer}.

As shown in Figure~\ref{fig:more_vis_attn_case}, we observe that although the model appears to correctly recognize the task (identifying the outlier based on shape features), it fails to execute it correctly on the query instance. Specifically, the attention maps reveal that the model focuses on an incorrect candidate object (\texttt{triangle}) in the query image rather than the true outlier (\texttt{square}). Consequently, despite the correct task formulation, this misdirected attention leads to an erroneous final prediction.


\section{Additional Discussion}
\textbf{Usage of Artifacts and AI Assistants.}
We utilized publicly available models and datasets from Hugging Face, strictly following their licenses for non-commercial research.
These models and datasets have been reviewed by their developers/creators to minimize the inclusion of personally identifiable information or offensive content and are widely adopted by the research community. We used AI tools to assist with language refinement during the writing process, the paper contains no AI-generated paragraphs. All material has been carefully reviewed to ensure accuracy and adherence to ethical standards.

\end{document}